\documentclass{article}
\usepackage{ijcai23}
\usepackage{times}
\usepackage{soul}
\usepackage{url}
\usepackage[hidelinks]{hyperref}
\usepackage[utf8]{inputenc}
\usepackage[small]{caption}
\usepackage{graphicx}
\usepackage{amsmath}
\usepackage{amssymb}
\usepackage{amsthm}
\usepackage{booktabs}
\usepackage{algorithm}
\usepackage{algorithmic}
\usepackage[switch]{lineno}
\usepackage{color}
\usepackage{multirow}
\usepackage{rotating}
\usepackage{bbding}

\title{OSP2B: One-Stage Point-to-Box Network for 3D Siamese Tracking}

\author{
Jiahao Nie$^1$ \and
Zhiwei He$^{1\dag}$\and
Yuxiang Yang$^1$\and
Zhengyi Bao$^1$\and
Mingyu Gao$^1$\and
Jing Zhang$^2$
\affiliations
$^1$School of Electronics and Information, Hangzhou Dianzi University, China\\
$^2$School of Computer Science, The University of Sydney, Australia
\emails
\{jhnie, zwhe, yyx, baozhengyi, mackgao\}@hdu.edu.cn,
jing.zhang1@sydney.edu.au
}

\begin{document}

\maketitle

\begin{abstract}
Two-stage point-to-box network acts as a critical role in the recent popular 3D Siamese tracking paradigm, which first generates proposals and then predicts corresponding proposal-wise scores. However, such a network suffers from tedious hyper-parameter tuning and task misalignment, limiting the tracking performance. Towards these concerns, we propose a simple yet effective one-stage point-to-box network for point cloud-based 3D single object tracking. It synchronizes 3D proposal generation and center-ness score prediction by a parallel predictor without tedious hyper-parameters. To guide a task-aligned score ranking of proposals, a center-aware focal loss is proposed to supervise the training of the center-ness branch, which enhances the network's discriminative ability to distinguish proposals of different quality. Besides, we design a binary target classifier to identify target-relevant points. By integrating the derived classification scores with the center-ness scores, the resulting network can effectively suppress interference proposals and further mitigate task misalignment. Finally, we present a novel one-stage Siamese tracker OSP2B equipped with the designed network. Extensive experiments on challenging benchmarks including KITTI and Waymo SOT Dataset show that our OSP2B achieves leading performance with a considerable real-time speed. Code will be available at https://github.com/haooozi/OSP2B.
\end{abstract}
\begin{figure}[t]
    \centering
    \includegraphics[width=\columnwidth]{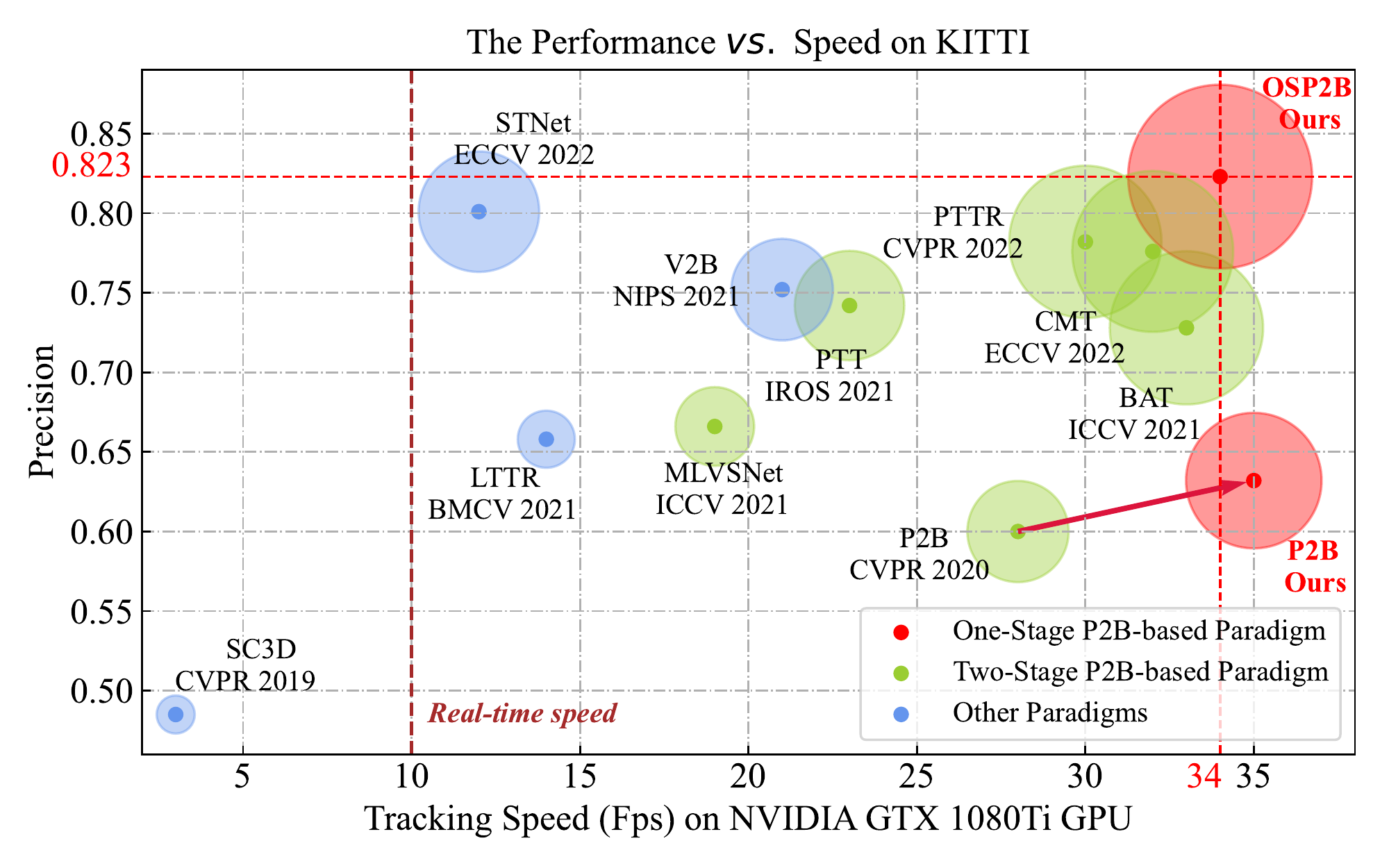}
    \caption{Comparison with SOTA methods on KITTI. We report the Precision performance with respect to tracking speed on a single NVIDIA GTX 1080Ti GPU. Circle size indicates the overall performance regarding Precision and tracking speed.}
    \label{fig1}
\end{figure}

\section{Introduction}
Single object tracking (SOT) is a fundamental task in computer vision and contributes to various applications, such as autonomous driving and mobile robotics \cite{things,survey}. Early tracking methods mainly focus on the 2D image domain. With the developments of LiDAR sensors, and considering that 3D point cloud data captured by LiDAR is more robust to adverse weather and illumination than RGB data, increasing efforts \cite{sc3d,3d-siamrpn,p2b,pttr} are devoted to point cloud-based tracking. In this paper, we study the problem of 3D SOT on LiDAR point clouds.

Recently, the Siamese network-based tracking paradigm has attracted remarkable attention. Current mainstream Siamese trackers such as P2B \cite{p2b}, BAT \cite{bat} and PTTR \cite{pttr} rely on a two-stage point-to-box network to \textit{learn the offsets of seed points towards object's center to generate 3D proposals (first stage), and then predict the highest scoring one as tracking box (second stage)}. Despite the great success, the two-stage design suffers from some inherent shortcomings: 1) A series of pre-defined hyper-parameters that require empirical and heuristic configurations. For example, tracking performance is sensitive to the number of proposals as reported in \cite{p2b}. 2) To predict proposal-wise scores, such a two-stage network defines proposals within a threshold distance from the object's center in 3D Euclidean space as positive samples for training. However, the nearest proposal has an equal contribution to the training as other positive ones, which exposes a task misalignment problem, i.e., the predicted highest-scoring proposal is not guaranteed to be the most accurate one.

Towards these concerns, we aim to offer a one-stage solution for 3D Siamese tracking. Inspired by one-stage 2D Siamese trackers \cite{siamcar,siamrpn}, the key to the one-stage design is to perform the proposal generation and proposal-wise score prediction simultaneously. Differently, in 2D tracking, the proposal-wise scores are predicted directly from image pixels without location information of the proposals, while for 3D point cloud tracking, seed points covering the surface of an object need to be offset towards the object's center to generate proposals for further score reasoning \cite{votenet}. Here, we argue that the seed point features that generate proposals are embedded with rich geometric cues, therefore it is feasible to synchronously predict the proposal-wise scores from these features. Additionally, we discover that sufficiently accurate proposals can usually be generated by offset learning (as verified in Section \ref{sec4.2}), so how to predict the most accurate one as a tracking result (i.e., task alignment) is significant for tracking task. To guide a task-aligned score ranking of proposals, distinguishing positive samples of different quality for training matters. Moreover, false-positive samples also need to be paid attention to avoid affecting the score ranking.

Motivated by the above analysis, we propose a simple yet effective \textbf{one-stage point-to-box network} that is free of tedious hyper-parameters, to improve the 3D Siamese tracking paradigm. Specifically, we design a parallel predictor and use the same seed point features to synchronize proposals generation and their center-ness scores prediction. Due to the 3D rigid object's constant size (width, height, and length) in point cloud sequences, the center-ness scores can effectively represent the accuracy of proposals. To distinguish positive samples of different quality to solve the task misalignment problem, we propose a center-aware focal loss to train the center-ness branch, in which a center-aware mask is devised to assign different loss weights for samples with regard to their proximity to the object's center. In addition, we develop a target classifier to classify foreground target points and background interference points. Leveraging classification scores, the center-ness scores can be refined to suppress false positive proposals to further alleviate the task misalignment. Finally, by integrating the proposed one-stage point-to-box network as the prediction head, a novel one-stage Siamese tracking method dubbed \textbf{OSP2B} is presented. As shown in Fig.~\ref{fig1}, OSP2B achieves state-of-the-art (SOTA) tracking performance, while running at a high speed of 34 frames per second (Fps). In particular, our one-stage point-to-box network outperforms the previous two-stage counterpart in both accuracy and efficiency, as clearly verified by P2B $v.s.$ P2B-ours in Fig.~\ref{fig1}. 

The main contributions of this paper are as follows: 
\begin{itemize}
    \item We propose a simple yet effective one-stage point-to-box network to improve the 3D Siamese paradigm, and present a novel OSP2B tracker for point cloud-based single object tracking. 
    \item We design a parallel predictor to synchronize 3D proposal generation and center-ness score prediction, avoiding tracking-sensitive hyper-parameters.
    \item We design a center-aware focal loss and a target classifier to guide a task-aligned score ranking of proposals, effectively addressing the task misalignment problem.
    \item Compared with SOTA methods, our OSP2B outperforms them in terms of both accuracy and efficiency on challenging benchmarks including KITTI and Waymo SOT Dataset.
\end{itemize}

\section{Related Work}
\subsection{2D Siamese Tracking}
Currently, Siamese network-based tracking methods \cite{siamfc,siamrpn,siamcar,transt} serve a dominant role in 2D visual object tracking. Generally, the Siamese tracking paradigm composed of two branches projects the target template and search images into an intermediate feature embedding space, and then fuses the template and search features by fusion modules such as cross-correlation \cite{ocean,siamla} and attention-based operators \cite{ttdimp,cia}. Subsequently, the fused features are further used to regress bounding boxes and calculate box-wise scores. Despite of the great success, it is non-trivial to extend 2D Siamese techniques to process 3D point cloud data.

\subsection{3D Siamese Tracking}
Early 3D tracking methods \cite{rgbd1,rgbd2,rgbd3,rgbd4,rgbd5} directly employ the 2D Siamese architecture to process RGB-D data with an additional depth channel. Recently, many efforts have been focused on tracking point cloud objects, as point cloud data is less sensitive to adverse weather than RGB-D data. As a pioneer, SC3D \cite{sc3d} proposes the first 3D Siamese tracker, but it is not an end-to-end framework and fails to run in real-time due to exhaustive 3D candidate boxes. To address these issues, P2B \cite{p2b} introduces a two-stage point-to-box network to perform proposal generation and proposal-wise score prediction for tracking, making a good balance between accuracy and speed. Inspired by this strong baseline, a series of follow-ups have been presented. BAT \cite{bat} and PTTR \cite{pttr} improve P2B by using different feature fusion modules to replace the point-wise correlation operator. PTT \cite{ptt}, MLVSNet \cite{mlvsnet} and GLT-T \cite{glt} propose to form a powerful feature presentation of seed points by designing more advanced structures. Although great progress has been made, these methods all follow the two-stage point-to-box network-based paradigm. By contrast, we offer a simpler and more effective one-stage design for 3D Siamese tracking.
\begin{figure*}[t]
    \centering
    \includegraphics[width=2.0\columnwidth]{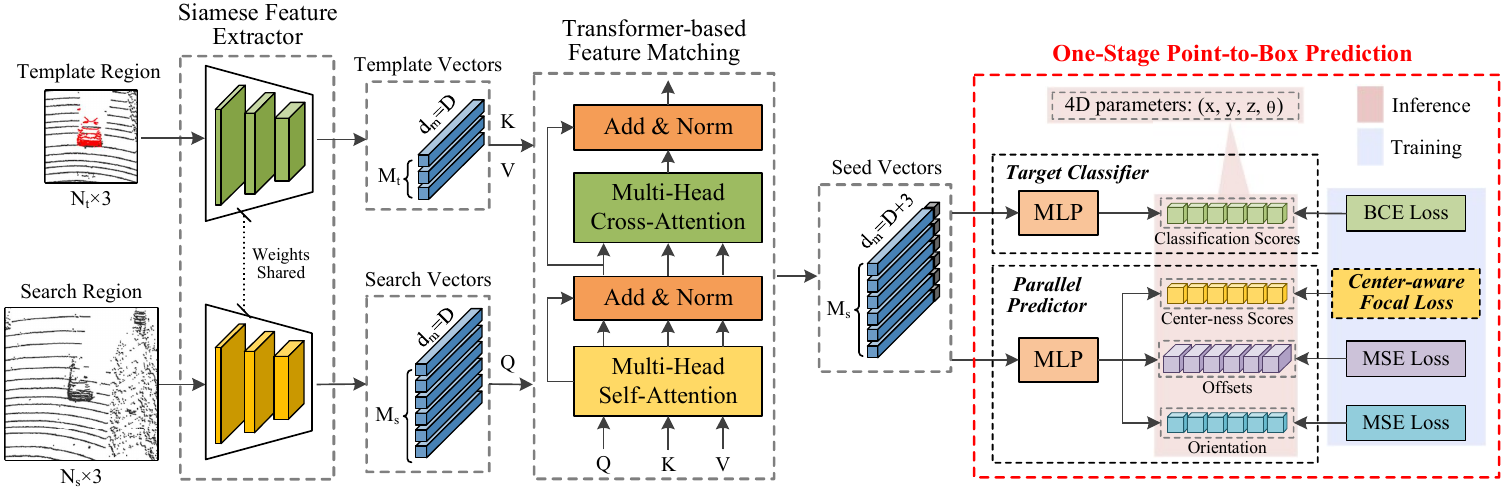}
    \caption{Overview of the proposed \textbf{OSP2B}. Given a template and search region, we first utilize a Siamese Feature Extractor to extract the point features, and fuse them with a Transformer-based feature matching module to output seed points. Finally, we apply the proposed One-Stage Point-to-Box Prediction network to predict a final 3D bounding box (BBox).}
    \label{fig2}
\end{figure*}

\subsection{Point-to-Box Network}
Point-to-box network is inspired by Hough voting \cite{votenet}, where a set of seed points are sampled to generate votes from their features, and the votes are targeted to reach the object's center. To apply the idea of Hough voting to 3D single object tracking, existing two-stage point-to-box network-based trackers first generate votes (i.e., proposals) by offsetting the seed points covering the surface of an object to its center, and form vote clusters via a shared PointNet \cite{pointnet}, including set-abstraction and propagation layers. With the vote clusters, a genetic point set learning network is then employed to predict scores and orientations of the votes, and refine the votes through secondary offset learning. In contrast, within the proposed one-stage point-to-box network, we synchronize proposal generation by a parallel predictor, and customize a center-aware focal loss and a target classifier to address the task misalignment problem.

\section{OSP2B: A Novel One-Stage Point-to-Box Network based 3D Siamese Tracker}
\subsection{Overall Architecture}
In a 3D scene, given a template target $\mathbf{P}^t=\{p_i^t\}_{i=1}^{N_t}$ cropped by the 3D bounding box (BBox) in the first frame, the tracking task aims to locate this target in search region $\mathbf{P}^s=\{p_i^t\}_{i=1}^{N_s}$ frame by frame. The 3D BBox is parameterized with a 7-dimensional vector, where $(x,y,z)$ and $(w,h,l)$ represent the center coordinate and size, while $\theta$ is the orientation. Since the object size is known in the first frame and keeps constant, only $(x,y,z,\theta)$ need to be predicted for tracking. To this end, we propose OSP2B, a novel one-stage paradigm for point cloud object tracing. As shown in Fig.~\ref{fig2}, our OSP2B consists of three key parts: Siamese feature extractor, transformer-based feature matching, and one-stage point-to-box prediction.

\noindent\textbf{Siamese Feature Extractor.} Following previous trackers \cite{p2b,bat}, we employ a modified PointNet++ \cite{pointnet++} as the backbone to subsample key points and extract their semantic features. More concretely, we remove the last task-relevant layers of PointNet++ and use it to encode multi-scale point features of template and search.

\noindent\textbf{Transformer-based Feature Matching.} Transformer-based feature matching is adopted to fuse the point features of the template and search to generate seed points. Similar to existing methods \cite{pttr,transt}, we first exploit a shared self-attention block~\cite{vit,xu2022vitpose,zhang2022vitaev2,zhang2022vsa} to enhance the feature representation of template and search, and then a cross-attention block is used to match them.

\noindent\textbf{One-Stage Point-to-Box Prediction.} Different from the existing two-stage point-to-box network (The detailed structure comparison is offered in the supplementary material), we propose a one-stage point-to-box network as prediction head to predict 4-dimensional BBox parameters $(x,y,z,\theta)$, as introduced in Section \ref{sec3.2}. We first design a parallel predictor to generate proposals and predict center-ness scores simultaneously. Then, a center-aware focal loss is devised to train the center-ness branch to distinguish positive samples with different quality. Finally, we also develop a target classifier to identify foreground target points and  background interference points to suppress false positive samples.

\subsection{One-Stage Point-to-Box Network}
\label{sec3.2}
\noindent\textbf{Parallel Predictor.} Given the seed point vectors $\mathbf{V}=\{v_i\in[f_i;p_i]\}_{i=1}^{M_s}$ as inputs, where $f_i$ and $p_i=(x_i,y_i,z_i)$ denote the semantic features and 3-dimensional coordinate, we design an offset branch and a center-ness branch to form a parallel predictor, which generates proposals and predicts center-ness scores, respectively. Meanwhile, an orientation branch is attached to predict the orientation of each proposal, as shown in the red box in Fig.~\ref{fig2}.

For proposal generation, the offset branch outputs a set of 3-dimensional vectors $\{d_i=(x_i^d,y_i^d,z_i^d)\}_{i=1}^{M_s}$ to predict the distances from the seed points $\{p_i\}_{i=1}^{M_s}$ to object's center. The proposals $\{p_i+d_i\}_{i=1}^{M_s}$ are obtained by applying offsets to the original coordinates of seed points. During training, let $(\tilde{x},\tilde{y},\tilde{z})$ represent the center of ground truth BBox, the target offset for the $i$-th seed point can be calculated by:
\begin{equation}
    t_i[0]=\tilde{x}-x_i,\quad t_i[1]=\tilde{y}-y_i,\quad  t_i[2]=\tilde{z}-z_i.
    \label{eq1}
\end{equation}
With $\{t_i\}_{i=1}^{M_s}$, we sample foreground target points and use $\rm {MSE}$ loss function to compute the offset loss as:
\begin{equation}
    \mathcal{L}_{off}=\frac{1}{M_s^{\prime}}\sum_{i=1}^{M_s^{\prime}}{\sum_{j=0}^{2}\|d_i[j]-t_i[j]\|^2},
    \label{eq2}
\end{equation}
where $M_s^{\prime}<M_s$ is the number of foreground target points.

For center-ness score prediction, the goal of the center-ness branch is to output scores that can denote the accuracy of proposals. To this end, a common solution typically calculates the distance between proposals and the object's center in 3D Euclidean space and collects the seed points corresponding to the proposals within a threshold as positive samples while treating other points as negative ones. However, this approach ignores the spatial distribution of proposals inside objects, especially slender objects, resulting in unbalanced positive and negative samples. We thereby propose shape-adaptive labels to balance the positive and negative sample sizes. In practice, instead of a sphere, the positive sample candidates are specified as in a rectangular cube that occupies by the object at scale $\tau$:
\begin{equation}
  \tilde{s}_i =
  \begin{cases}
  1 & \text{if } (x_i+x_i^d,y_i+y_i^d,z_i+z_i^d)\in \mathbb{R}^{\tau(\tilde{w},\tilde{h},\tilde{l})}\\
  0 & \text{otherwise}, \\
  \end{cases} 
  \label{eq3}
\end{equation}
where $\tilde{s}_i$ is the label for seed point $p_i$, $\tilde{w}$, $\tilde{h}$ and $\tilde{l}$ denotes the width, height, and length of ground truth BBox, respectively. The center-ness loss is computed by a center-aware focal loss function, which is detailed below.

To characterize the orientations of cubic objects in 3D space, we train the orientation branch by:
\begin{equation}
    \mathcal{L}_{ori}=\frac{1}{M_s^{\prime}}\sum_{i=1}^{M_s^{\prime}}{\|\theta_i-\tilde{\theta}\|^2},
    \label{eq4}
\end{equation}
where $\theta_i$ and $\tilde{\theta}$ denote the predicted orientation of each proposal and the orientation of ground truth BBox.

\noindent\textbf{Center-aware Focal Loss.} In fact, the center-ness scores should relate to the proximity of proposals to the object's center. Therefore, a training strategy that can distinguish positive samples of different proximity is desired to train the center-ness branch. To achieve this, we customize a center-aware focal loss. Specifically, a center-aware point mask is defined as:
\begin{equation}
    mask_i=\sqrt[3]{\frac{{\rm min}(l,r)}{{\rm max}(l,r)}\times \frac{{\rm min}(t,b)}{{\rm max}(t,b)}\times \frac{{\rm min}(f,k)}{{\rm max}(f,k)}},
    \label{eq5}
\end{equation}
where $l$, $r$, $t$, $b$, $f$ and $k$ represent the distance of proposal point $p_i+d_i$ to the left, right, top, bottom, front and back surfaces of ground truth BBox, respectively. In this way, the samples closer to the object's center tend to have higher mask scores. Here, we consider point mask scores as loss weights for different positive samples, thereby incorporating center-aware geometry prior into the model training. To this end, the center-aware focal loss is formulated as:
\begin{equation}
  {\rm CAFL}(s_i,\tilde{s}_i) =
  \begin{cases}
  -\alpha(1-s_i)^{\gamma}(1+mask_i){\rm log}(s_i) & \tilde{s}_i=1\\
  -\beta(s_i)^{\gamma}{\rm log}(1-s_i) & \tilde{s}_i=0, \\
  \end{cases} 
  \label{eq6}
\end{equation}
where $s_i$ is the predicted center-ness score of $i$-th proposal. Following \cite{focalloss}, we empirically set the three hyper-parameters $\alpha$=2, $\beta$=1, and $\gamma$=2 to balance the loss weights for positive (easy) and negative (hard) samples. The center-ness loss is:
\begin{equation}
     \mathcal{L}_{cen}=\frac{1}{M_s}\sum_{i=1}^{M_s}{{\rm CAFL}(s_i,\tilde{s}_i)}.
    \label{eq7}
\end{equation}

\noindent\textbf{Target Classifier.} Since background interference points are not supervised in offset training, these points might be sampled as positive samples for center-ness training, resulting in proposals far from the center being predicted high scores. To make the tracker resist interference, a target classifier is also devised. We take the foreground seed points inside the 3D object BBox as positive samples and the other points as negative ones. Using the vanilla binary cross-entropy loss function, the classifier loss can be defined as:
\begin{equation}
     \mathcal{L}_{cla}=-\frac{1}{M_s}\sum_{i=1}^{M_s}{(\tilde{c}_i{\rm log}(c_i)+(1-\tilde{c}_i){\rm log}(c_i))},
    \label{eq8}
\end{equation}
where $c_i$ is the predicted classification score, $\tilde{c}_i=$1 or 0 is the corresponding label.

Considering the different contributions of seed points to offset learning, we also use classification scores to supervise the offset branch training except for adjusting center-ness scores. The offset loss in Eq. \ref{eq2} is further refined as:
\begin{equation}
    \mathcal{L}_{off}=\frac{1}{M_s^{\prime}}\sum_{i=1}^{M_s^{\prime}}{(\sum_{j=0}^{2}\|d_i[j]-t_i[j]\|^2)(1+c_i)},
    \label{eq9}
\end{equation}
where $c_i$ allows the model to focus more on the points inside the object, facilitating offset learning and consequently generating better proposals.

\subsection{Implementation}
\noindent\textbf{Model Inputs.} During training, we sample paired samples, i.e., template and search regions from the same point cloud sequences. The template region is formed by merging the points inside the ground truth BBoxes of $(t-1)$-th frame and $1$-st frame. For the search region, since the target shifts a little between consecutive frames, only a prior region where the target may appear is required. We thereby enlarge the ground truth BBox of $t$-th frame by 2 meters and crop the points within this enlarged area. To enhance the robustness of the model, we randomly impose small shifts to the BBoxes along the $x$, $y$, and $z$ axes in the training phase. 

During inference, the ground truth BBox of $1$-st frame is given, but the ground truth BBox of subsequent frames is unknown. Therefore, when generating the template region and search region of $t$-th frame, the ground truth BBox of $(t-1)$-th and $t$-th frames mentioned above is replaced by the BBox of $(t-1)$-th frame predicted by the model.

\noindent\textbf{Model Details.} We randomly sample $N_t=512$ and $N_s=1024$ points for the template region and search region, respectively. Then a modified PointNet++ \cite{pointnet++} with 3 set-abstraction layers is adopted as the Siamese backbone, to obtain the semantic features of key points, where $M_t=64$, $M_s=128$ and $D=256$. In the proposed one-stage point-to-box prediction head, the hidden layers are built by a 2-layer MLP that has a constant channel dimension.

\noindent\textbf{Training.} Our OSP2B can be trained in an end-to-end manner. With the above losses of four branches in the one-stage point-to-box network, the total loss is defined as:
\begin{equation}
    \mathcal{L}=\lambda_1\mathcal{L}_{off}+\lambda_2\mathcal{L}_{ori}+\lambda_3\mathcal{L}_{cen}+\lambda_4\mathcal{L}_{cla},
    \label{eq10}
\end{equation}
where $\lambda_1$, $\lambda_2$, $\lambda_3$ and $\lambda_4$ are all set to 1 to balance these losses. We use Adam optimizer to train the OSP2B model on 4 NVIDIA GTX 1080Ti GPUs for 160 epochs. The initial learning rate is set to 0.001 and decreased by a linear decay factor of 0.2 every 40 epochs. We provide the implementation code in the supplement.

\noindent\textbf{Inference.} To predict 4-dimensional vector $(x,y,z,\theta)$ of the target object frame by frame, OSP2B produces a 6-dimensional vector $(x_i^d,y_i^d,z_i^d,\theta_i,s_i,c_i)$ for each seed point $p_i=(x_i,y_i,z_i)$. The inference phase can be formulated as:
\begin{equation}
     I = \underset{i}{\rm argmax}\{s_i \times c_i \},
    \label{eq11}
\end{equation}
where $I$ is the index of the best proposal predicted by the OSP2B model, and the 4-dimensional vector $(x,y,z,\theta)$ of final tracking BBox is calculated by:
\begin{equation}
    x=x_I+x_I^d, \quad y=y_I+y_I^d, \quad
    z=z_I+z_I^d, \quad \theta=\theta_I.
    \label{eq12}
\end{equation}

\section{Experiments}
\begin{figure}[t]
    \centering
    \includegraphics[width=1.0\columnwidth]{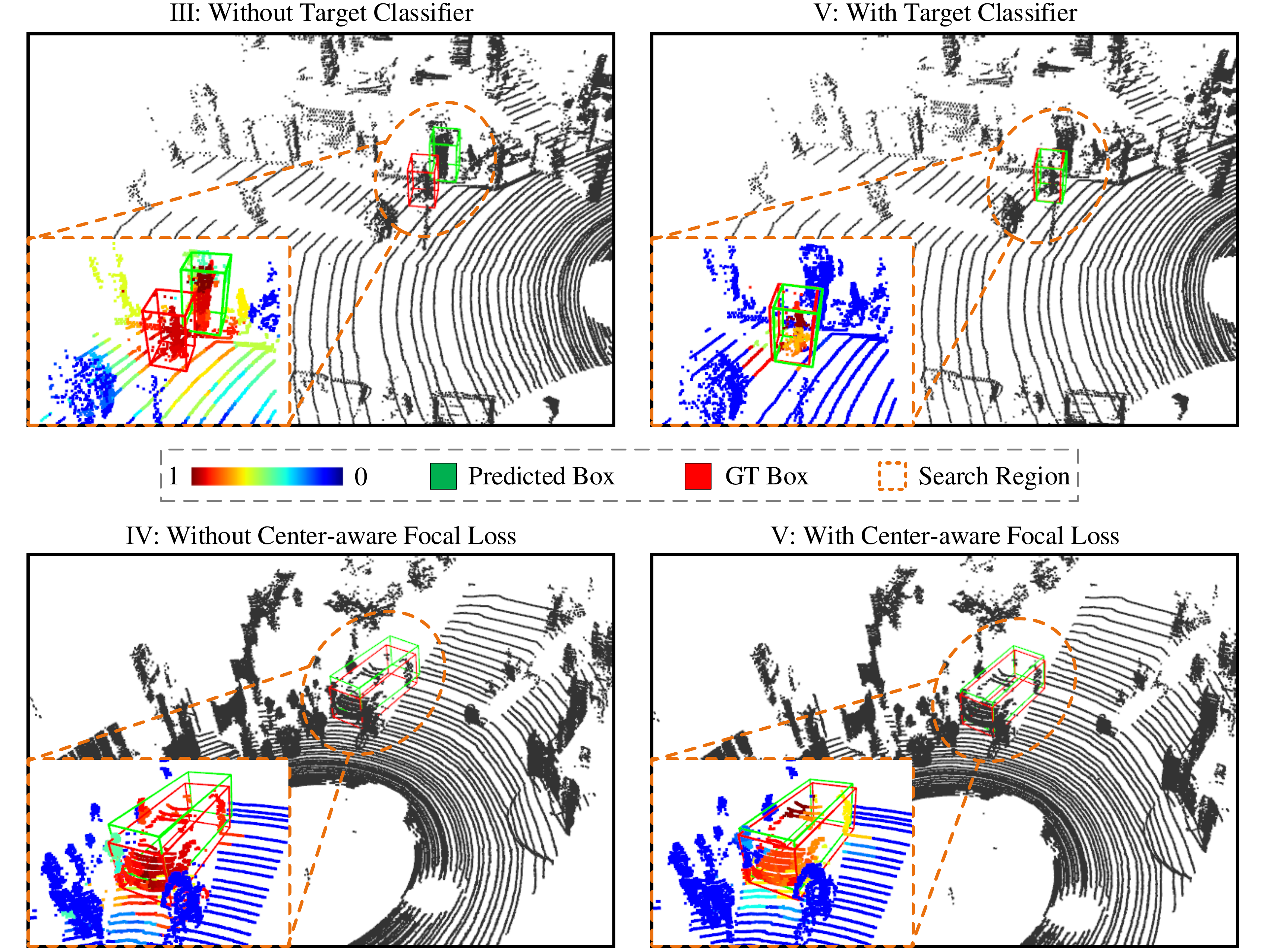} 
    \caption{Visualization of tracking results without or with target classifier (top row) and center-aware focal loss (bottom row), respectively. The final scores of proposals range from 0 to 1.}
    \label{fig3}
\end{figure}
\subsection{Experimental Settings}
\label{sec4.1}
\noindent\textbf{Datasets.} To evaluate our model, we conduct comprehensive experiments, including the ablation study on KITTI (Section \ref{sec4.2}) and comparison with SOTA methods on KITTI and Waymo SOT Dataset (Section \ref{sec4.3}). KITTI \cite{kitti} contains 21 training LiDAR sequences and 29 test LiDAR sequences. Due to the labels of test data is not open, we split the training set for training, validation, and testing, following the previous works \cite{p2b}. Waymo SOT Dataset \cite{pttr} is a more challenging and large-scale dataset, recently collected from the raw Waymo data \cite{waymo}. To be a fair comparison, we perform training and testing based on the method described in \cite{pttr}. 

\noindent\textbf{Evaluation Metrics.} Following the common practice, we calculate \textit{Success} and \textit{Precision} metrics by One Pass Evaluation (OPE) \cite{otb} to report tracking performance. \textit{Success} measures the intersection over union (IOU) between predicted BBox and ground truth BBox. \textit{Precision} measures the distance between the centers of two BBoxes.

\subsection{Ablation Study}
\label{sec4.2}
In this section, we conduct a series of ablation studies, including qualitative and quantitative analyses to validate the effectiveness of the proposed one-stage point-to-box network. As previous works \cite{sc3d,p2b,bat}, all ablated experiments are carried out on the Car category from the KITTI dataset.

\noindent\textbf{Model Component.} To investigate the contributions of the designed components: parallel predictor, center-aware focal loss, and target classifier to the tracking performance, a component-wise ablation experiment is conducted. We report the results of OSP2B using different prediction heads, as shown in Table~\ref{table1}. Compared to the two-stage point-to-box network ($ \rm{\uppercase\expandafter{\romannumeral1}}$), the proposed one-stage point-to-box network with only parallel predictor ($ \rm{\uppercase\expandafter{\romannumeral2}}$) outperforms it by 1.2\% and 1.9\% in terms of Success and Precision, respectively, proving the effectiveness of our one-stage design in synchronizing 3D proposals generation and center-ness scores prediction. For ($ \rm{\uppercase\expandafter{\romannumeral3}}$), the center-aware focal loss distinguishes positive samples with different quality in training, which increases the probability of the most accurate proposal being selected, and thus a significant performance improvement of 2.5\% and 4.3\% in Success and Precision is achieved. For ($ \rm{\uppercase\expandafter{\romannumeral4}}$), the target classifier suppresses the interference points and guides the network to predict more accurate scores, also leading to better performance. In addition, we visualize the tracking results of ($ \rm{\uppercase\expandafter{\romannumeral3}}$ $v.s.$ $\rm{\uppercase\expandafter{\romannumeral5}}$) and ($ \rm{\uppercase\expandafter{\romannumeral4}}$ $v.s.$ $ \rm{\uppercase\expandafter{\romannumeral5}}$) in Fig.~\ref{fig3} to intuitively demonstrate the effectiveness of the center-aware focal loss and the target classifier. As can be seen, the background interference points are given low scores by using the target classifier (top row), and the points closer to the object's center have higher scores than those far away by using the center-aware focal loss (bottom row). When combining all components ($ \rm{\uppercase\expandafter{\romannumeral5}}$), we achieve the best Success and Precision of 67.5\% and 82.3\%. 
\begin{table}[!t]
    \centering
    \resizebox{0.9\columnwidth}{!}{
    \begin{tabular}{c|ccc|cc}
      \toprule[0.5mm]
   &Parallel & Center-aware & Target & \multirow{2}{*}{\textit{Success}} & \multirow{2}{*}{\textit{Precision}}\\
     &Predictor & Focal Loss & Classifier \\
      \midrule
      \midrule
     \uppercase\expandafter{\romannumeral1} & & & &  64.3 & 76.4 \\
     \uppercase\expandafter{\romannumeral2}& \Checkmark &&& 65.5$_{\uparrow1.2}$ & 78.3$_{\uparrow1.9}$ \\
     \midrule
     \uppercase\expandafter{\romannumeral3}&\Checkmark&\Checkmark&& 66.8$_{\uparrow2.5}$ & 80.7$_{\uparrow4.3}$ \\
       \uppercase\expandafter{\romannumeral4} &\Checkmark& & \Checkmark&  66.1$_{\uparrow1.8}$ & 79.6$_{\uparrow3.2}$ \\
       \uppercase\expandafter{\romannumeral5} &\Checkmark& \Checkmark& \Checkmark& \textbf{67.5}$_{\uparrow3.2}$ & \textbf{82.3}$_{\uparrow5.9}$ \\
      \bottomrule[0.5mm]
    \end{tabular}}
     \caption{Ablation study of model components on Car category from KITTI. \textbf{Bold} denotes the best result.}
    \label{table1}
\end{table}

\begin{figure}[t]
  \centering
  \includegraphics[width=1.67in]{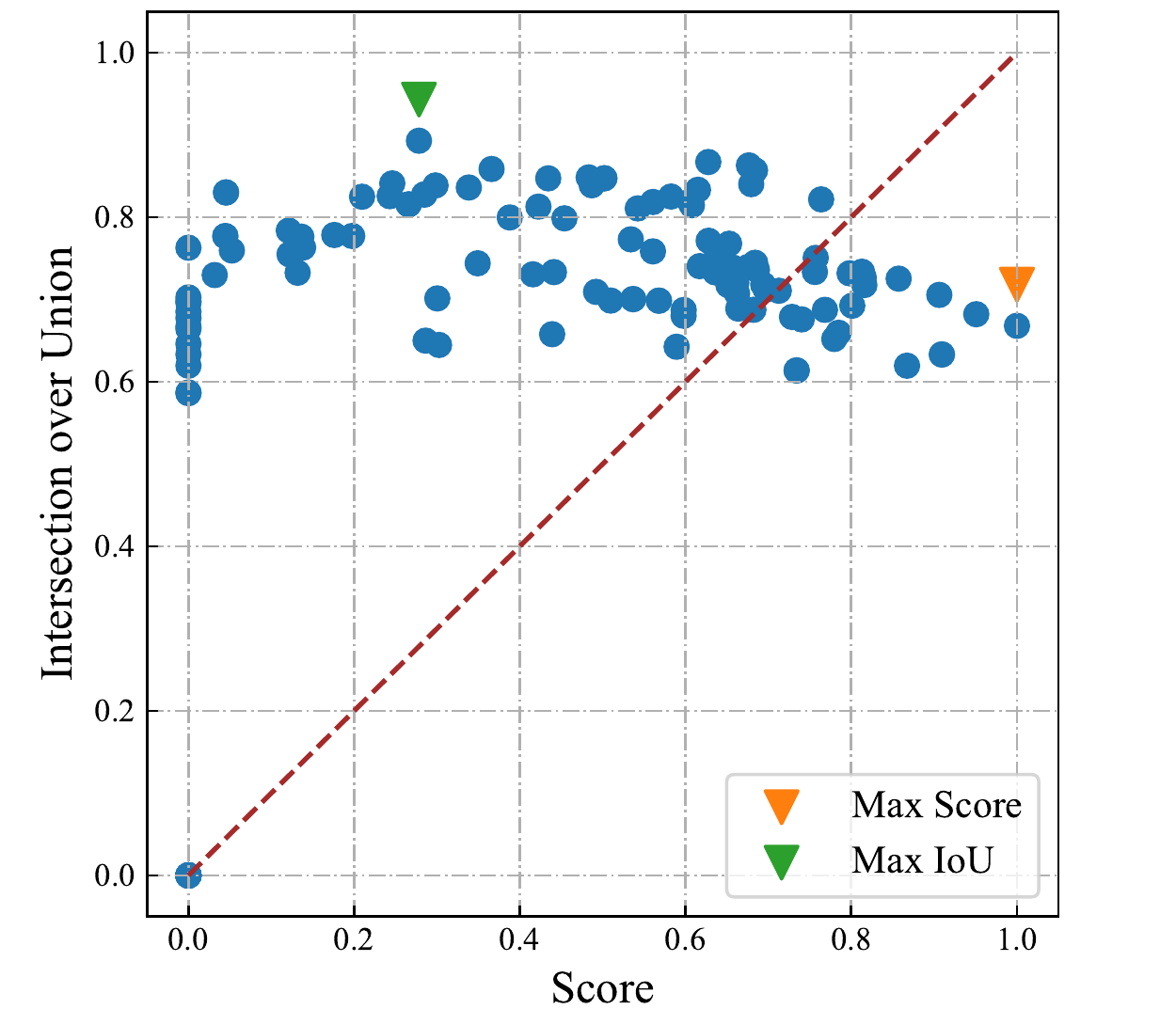}
  \includegraphics[width=1.67in]{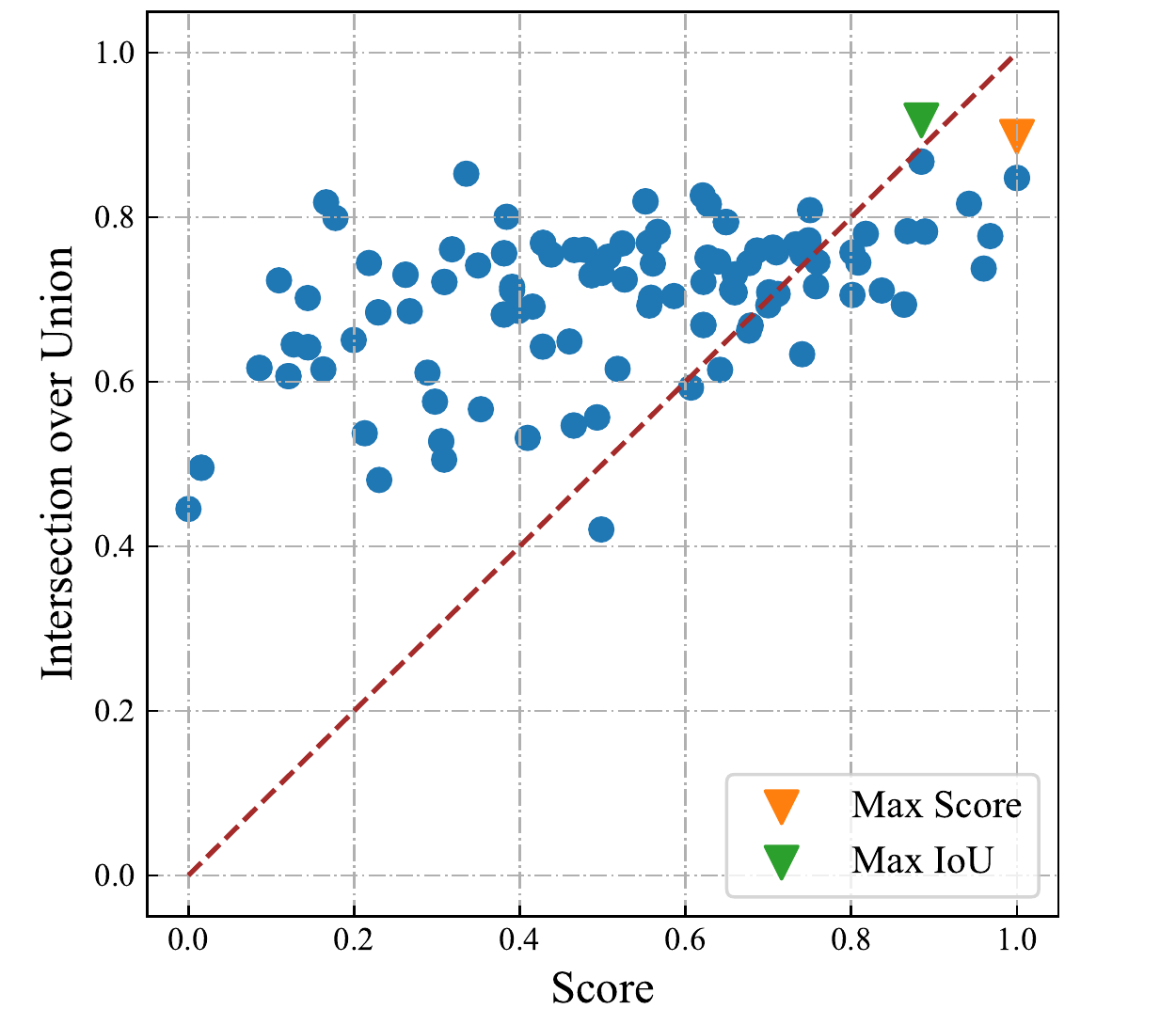}
  \caption{Visualization of the IoU $v.s.$ Score. Left: two-stage point-to-box network. Right: one-stage point-to-box network (ours).}
  \label{fig4}
\end{figure}

\noindent\textbf{Effectiveness of Task Alignment.} Here, we present the effectiveness of task alignment by visualizing IoU $v.s.$ Score of 3D proposals. From Fig.~\ref{fig4}, we have two observations: 1) The overall accuracy of the proposals generated by our proposed one-stage prediction head is slightly lower than that of the two-stage head, but the accuracy of the best one is comparable; 2) In our one-stage head, the better proposals are more likely to be predicted as the tracking results. In other words, the predicted scores are better aligned with the quality of the proposals. The two observations imply that predicting the best proposal as the tracking box contributes more remarkably to the tracker, compared to refining the proposal using the two-stage design. Owing to the proposed center-ness focal loss and target classifier, the task misalignment problem is effectively alleviated in our one-stage prediction head.
\begin{table}[!t]
    \centering
    \resizebox{0.5\columnwidth}{!}{
    \begin{tabular}{c|cc}
      \toprule[0.5mm]
      Scale Rate $\tau$ & \textit{Success}& \textit{Precision}\\
      \midrule
      \midrule
      0.3 & 64.7 & 76.6 \\
      0.4 & 67.3 & 81.9 \\
      0.5 & \textbf{67.5} & \textbf{82.3} \\
      0.6 & 66.9 & 80.4 \\
      0.7 & 65.7 & 78.6 \\
      \bottomrule[0.5mm]
    \end{tabular}}
    \caption{Influence of different scale rate $\tau$ on Car category from KITTI. \textbf{Bold} denotes the best result.}
    \label{table2}
\end{table}

\noindent\textbf{Scale Rate.} Scale rate $\tau$ is a significant hyper-parameter in our proposed one-stage point-to-box network. Too small a value will result in insufficient positive samples for training, while too large a value will distract the model and affect its discriminative ability. Therefore, we conduct an experiment to determine the optimal value of $\tau$. As reported in Table \ref{table2}, when $\tau=0.5$, the best \textit{Success} and \textit{Precision} values are obtained. So we set $\tau$ to 0.5 for all experiments if not specified.
\begin{figure*}[t]
    \centering
    \includegraphics[width=0.95\linewidth]{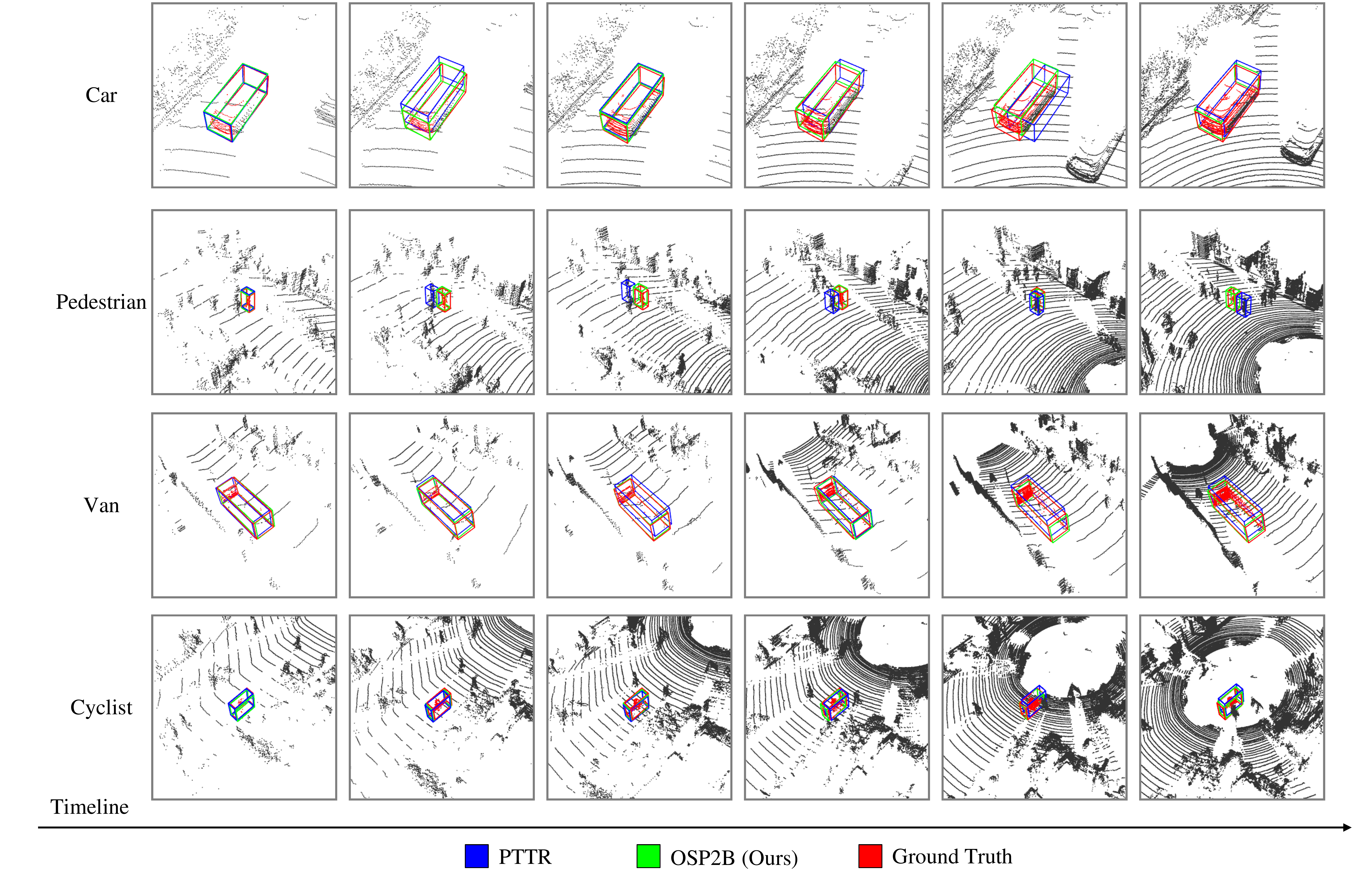}
    \caption{Visual tracking results of our OSP2B and PTTR on the point cloud sequences of Car, Pedestrian, Van, and Cyclist categories. The foreground points within the ground truth BBoxes are colored in red.}
    \label{fig5}
\end{figure*}

\subsection{Comparison with SOTA Methods}
\label{sec4.3}
\noindent\textbf{Results on KITTI.} We compare our OSP2B with 9 SOTA methods on four categories from KITTI \cite{kitti}. The results are presented in Table \ref{table3}. OSP2B achieves state-of-the-art performance in all categories. Especially for Pedestrian, an impressive performance advantage over other comparison methods is exhibited. Besides, we obtain the best mean Precision of 82.3\%, which suggests that our method is able to accurately predict object centers. Compared to the most recent method STNet \cite{stnet}, the proposed method not only shows competitive performance in Car, Van, and Cyclist categories but also surpasses it in the Pedestrian category by a large margin. Moreover, OSP2B runs nearly three times faster than STNet on the same experimental setting (34 Fps $v.s.$ 12 Fps). Notably, PTTR \cite{pttr} has a similar architecture to our OSP2B, except for the prediction head. Compared to it, OSP2B shows significant performance improvements in all categories, such as the Car category (\textit{Success}: 65.2\%$\to$67.5\%; \textit{Precision}: 77.4\%$\to$82.3\%), which manifests the superiority of our one-stage design. In addition, to intuitively compare the proposed one-stage point-to-box network and the two-stage point-to-box network used in PTTR, we also visualize their results from the four categories. As shown in Fig. \ref{fig5}, our method can track different categories of point cloud objects more accurately and robustly.
\begin{table}[!t]
    \centering
    \resizebox{\columnwidth}{!}{
    \begin{tabular}{cc|ccccc}
      \toprule[0.5mm]
      & Category & Car & Ped &  Van & Cyc & Mean \\
      & Frame Number & 6,424 & 6,088 & 1,248 & 308 & 14,068 \\
      \midrule
      \midrule
      \multirow{11}{*}{\rotatebox{90}{\textit{Success}}}
      &SC3D \cite{sc3d}& 41.3  & 18.2 & 40.4  & 41.5 &31.2  \\
      & P2B \cite{p2b}& 56.2  & 28.7  & 40.8  & 32.1 &42.4  \\
       &MLVSNet \cite{mlvsnet}& 56.0  & 34.1 & 52.0  & 34.3 & 45.7  \\
       &LTTR \cite{lttr}& 65.0  & 33.2 & 35.8  & 66.2  &48.7 \\
       &BAT \cite{bat}& 60.5  & 42.1 & 52.4  &33.7 & 51.2  \\
      &PTT \cite{ptt}& 67.8  & 44.9  & 43.6  & 37.2  & 55.1 \\
      &V2B \cite{v2b}& 70.5  & 48.3 & 50.1 & 40.8  &58.4  \\
      &PTTR \cite{pttr}& 65.2  & \underline{50.9} & 52.5  & 65.1 & 57.9 \\
      &CMT \cite{cmt}& 70.5  & 49.1 & 54.1  & 55.1 & 59.4 \\
       &STNet \cite{stnet}& \textbf{72.1} & 49.9 & \textbf{58.0} &  \textbf{73.5} & \textbf{61.3} \\
      \cmidrule{2-7}
      & OSP2B (ours) & 67.5 & \textbf{53.6} & \underline{56.3}  & \underline{65.6}  & \underline{60.5} \\
      \midrule
      \midrule
      \multirow{11}{*}{\rotatebox{90}{\textit{Precision}}}
      &SC3D \cite{sc3d}& 57.9 &  37.8 & 47.0 &  70.4 & 48.5 \\
      & P2B \cite{p2b}&  72.8 &  49.6 &  48.4 &  44.7 & 60.0 \\
       &MLVSNet \cite{mlvsnet}&  74.0 &  61.1 &  61.4 & 44.5 &  66.6 \\
      &LTTR \cite{lttr}& 77.1 & 56.8 &  48.4 & 89.9 & 65.8 \\
      &BAT \cite{bat}&  77.7 & 70.1 & 67.0 &  45.4 &  72.8 \\
      &PTT \cite{ptt}& 81.8 & 72.0 &  52.5 &  47.3 &  74.2 \\
      &V2B \cite{v2b}&  81.3 &  73.5 & 58.0 &  49.7 & 75.2 \\
      &PTTR \cite{pttr}& 77.4 & \underline{81.6} & 61.8 & 90.5& 78.2\\
       &CMT \cite{cmt}& 81.9  & 75.5 & 64.1  & 82.4 & 77.6 \\
      &STNet \cite{stnet}& \textbf{84.0} &  77.2 & \textbf{70.6} &  \textbf{93.7} &  \underline{80.1} \\
      \cmidrule{2-7}
       & OSP2B (ours) & \underline{82.3} & \textbf{85.1} & \underline{66.2} & \underline{90.5} & \textbf{82.3} \\
      \bottomrule[0.5mm]
    \end{tabular}}
     \caption{Comparison on Car, Pedestrian, Van, and Cyclist categories from KITTI benchmark. \textbf{Bold} and \underline{underline} denote the best result and the second-best one, respectively.}
    \label{table3}
\end{table}
\begin{table}[!t]
    \centering
    \resizebox{0.95\columnwidth}{!}{
    \begin{tabular}{cc|cccc}
      \toprule[0.5mm]
      &Category & Veh & Ped & Cyc & Mean\\
      &Frame Number & 53,377 & 27,308 & 5,374& 86,095 \\
      \midrule
      \midrule
      \multirow{4}{*}{\rotatebox{90}{\textit{Success}}}
      &SC3D \cite{sc3d}& 46.5  & 26.4 & 26.5 & 33.1  \\
      &P2B \cite{p2b}& 55.7  & 35.3 & 30.7 &40.6  \\
      &PTTR \cite{pttr}& \underline{58.7}  & \textbf{49.0}  & \textbf{43.3} & \textbf{50.3} \\
      \cmidrule{2-6}
      &OSP2B (ours) & \textbf{59.2} & \underline{46.6} & \underline{43.0}  &\underline{49.6} \\
      \midrule
      \midrule
      \multirow{4}{*}{\rotatebox{90}{\textit{Precision}}}
      &SC3D \cite{sc3d}& 52.7  & 37.8 & 37.6 & 42.7  \\
      &P2B \cite{p2b}& 62.2  & 54.9 & 44.5 &53.9  \\
      &PTTR \cite{pttr}& \underline{65.2} & \textbf{69.1}  & \underline{60.4} & \underline{64.9} \\
      \cmidrule{2-6}
      &OSP2B (ours) & \textbf{67.3} & \underline{67.4} & \textbf{62.5}  &\textbf{65.7} \\
      \bottomrule[0.5mm]
    \end{tabular}}
    \caption{Comparison on Vehicle, Pedestrian, and Cyclist categories from Waymo SOT Dataset benchmark. \textbf{Bold} and \underline{underline} denote the best result and the second-best one, respectively.}
     \label{table4}
\end{table}

\noindent\textbf{Results on Waymo SOT Dataset.} To further evaluate the proposed OSP2B method, we also conduct comparison experiments on the large-scale dataset Waymo SOT Dataset \cite{pttr}. We select SC3D \cite{sc3d}, P2B \cite{p2b} and PTTR \cite{pttr}, which have reported performance on this dataset as comparison methods. As presented in Table \ref{table4}, OSP2B achieves state-of-the-art performance in all categories, demonstrating that our one-stage method not only performs well on the small-scale dataset but also delivers satisfactory results on the large-scale dataset. Besides, since the Waymo SOT Dataset benchmark contains a wide range of complex real-world scenes, the superior performance of the proposed method indicates that it has great potential for practical applications.

\noindent\textbf{Inference Speed.} In addition to tracking accuracy comparisons, we also compare the inference speed of our OSP2B with SOTA methods. For a fair comparison, the average running time of each tracker is calculated on all test frames in the Car category from KITTI. OSP2B runs at 34 Fps on a single NVIDIA 1080Ti GPU, including 7.6 ms for processing point cloud, 21.1 ms for network forward propagation, and 0.8 ms for post-processing. The running speeds of other methods under the same workstation are reported in Table \ref{table5}. Thanks to the efficient one-stage point-to-box prediction head, our OSP2B achieves the fastest inference speed.
\begin{table}[!t]
    \centering
    \resizebox{0.95\columnwidth}{!}{
    \begin{tabular}{c|c||c|c}
      \toprule[0.5mm]
      Method & Fps & Method &  Fps \\
      \midrule
      \midrule
       SC3D \cite{sc3d} & 3 &  PTT \cite{ptt} & 23 \\
       P2B \cite{p2b} & 28 &V2B \cite{v2b} & 21 \\
       MLVSNet \cite{mlvsnet} & 19 &PTTR \cite{pttr} & 30 \\
       LTTR \cite{lttr} &   14 & STNet \cite{stnet} & 12 \\
       BAT \cite{bat} & 33  & OSP2B (ours) & \textbf{34} \\
      \bottomrule[0.5mm]
    \end{tabular}}
    \caption{Speed comparison on all test frames in the Car category from KITTI. \textbf{Bold} denotes the best result.}
    \label{table5}
\end{table}

\section{Limitation Discussion}
\begin{figure}[t]
    \centering
    \includegraphics[width=0.95\columnwidth]{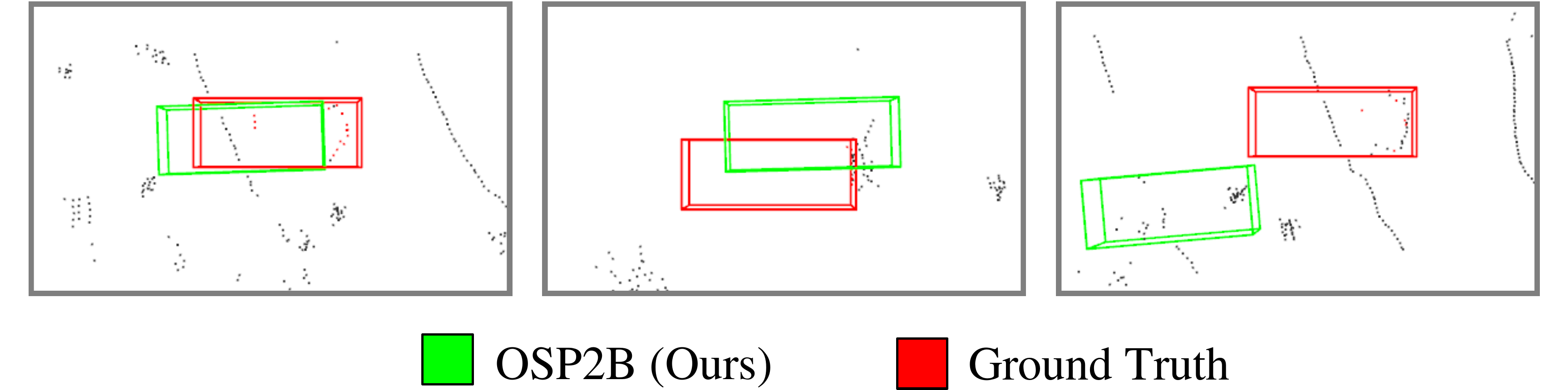}
    \caption{Tracking failure cases of our OSP2B on extremely sparse point cloud scene.}
    \label{fig6}
\end{figure}

We show the tracking failure cases of our OSP2B in Fig.~\ref{fig6}. It can be seen that OSP2B is not ready to handle extremely sparse point cloud scenes. This is mainly owing to the inability of the model to infer offsets from a small number of points, and thus causing 3D proposals to drift from the object center. One possible solution is to use the point cloud completion technique to obtain the model inputs with rich point cloud information.

\section{Conclusion}
In this paper, we revisit the 3D single object tracking on LiDAR point clouds, and propose to boost the Siamese paradigm with a novel one-stage point-to-box network, which is demonstrated to be superior over the two-stage counterpart by comprehensive experiments and analysis. By integrating this network as a prediction head, we develop a one-stage Siamese tracking method OSP2B, which can track point cloud objects in a one-stage manner and effectively address the task misalignment problem. Benefiting from the one-stage design, our OSP2B significantly outperforms previous SOTA trackers in terms of both accuracy and efficiency on challenging datasets. We hope OSP2B could serve as a one-stage baseline method and inspire future research on accurate and efficient 3D single object trackers.
\bibliographystyle{named}
\bibliography{ijcai23}
\clearpage

\section*{Supplementary Material}
\begin{figure*}[t]
    \centering
    \includegraphics[width=2.0\columnwidth]{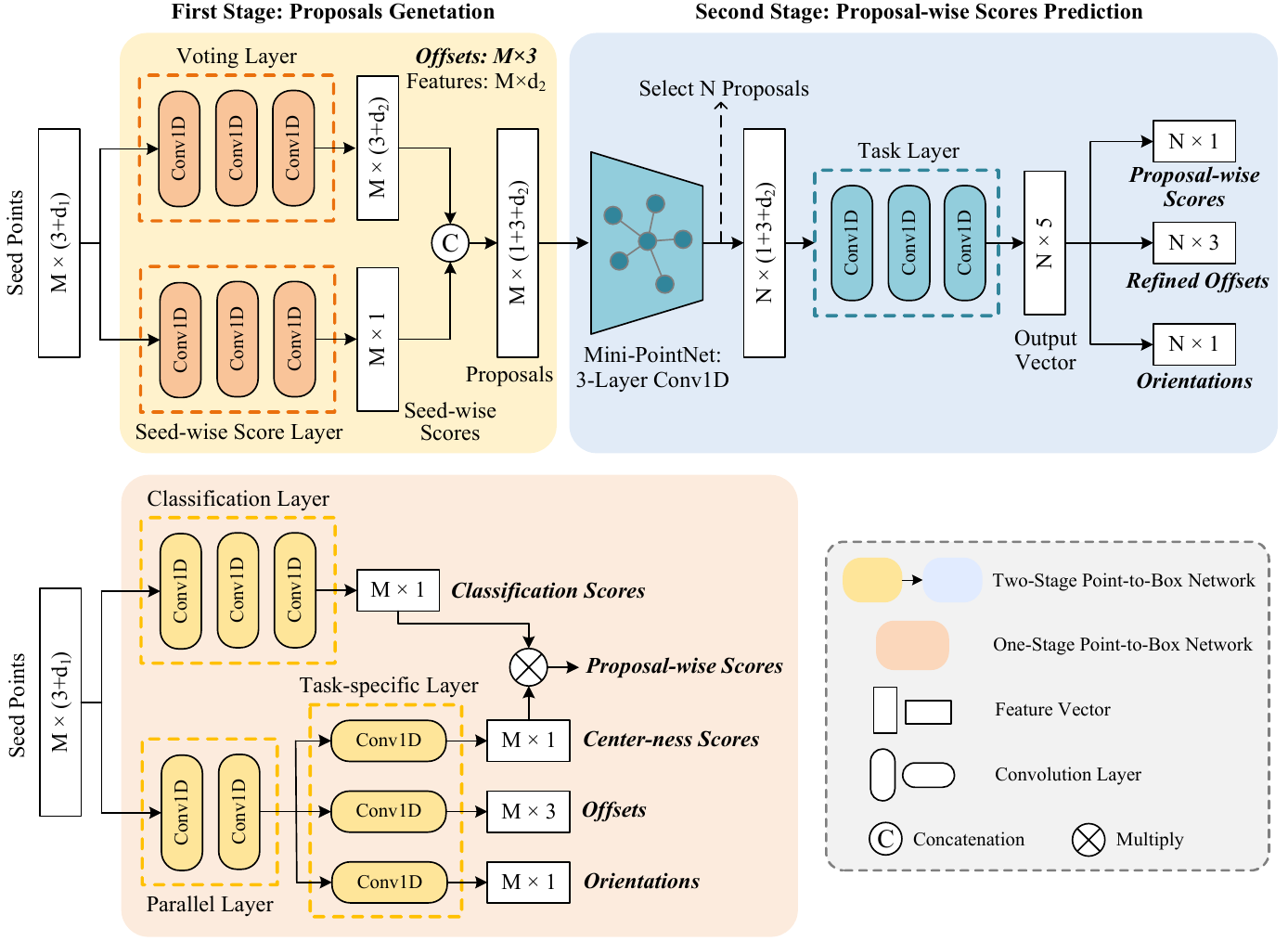}
    \caption{Comparison of different predicted head in Siamese tracking paradigm. Top row: previous two-stage point-to-box network is first proposed in P2B and used as a common prediction head in existing methods. Bottom row: our novel one-stage point-to-box network is presented.}
    \label{fig7}
\end{figure*}
\subsection*{A. Overview}
In this supplementary material, to further prove our work, we use P2B \cite{p2b} as baseline and provide a series of comparison between the proposed one-stage point-to-box network and previous two-stage point-to-box network, as follows:
\begin{itemize}
    \item The comparison of structure details.
    \item The comparison of performance under a variety of experimental setting.
        \begin{itemize}
        \item Tracking-Sensitive Parameters.
        \item Robustness to Sparse Point Clouds. 
        \item Ways for Template Region Generation. 
        \item Ways for Search Region Generation. 
        \item Quantitative Results.
        \item Model Parameters, Flops and Speed.
        \end{itemize}
\end{itemize}

\subsection*{B. Structure Comparison}
\label{subsection2}
Fig. \ref{fig7} shows the detailed structure comparison. For two-stage point-to-box network, it first learns offset features (3-dimensional coordinates and $d_2$-dimensional semantic features) by a voting layer composed of 3-layer 1D convolution. Meanwhile, a seed-wise score layer is used to predict seed-wise scores, which are concatenated with the offset features to produce $M$ proposals. After that, $N$ proposals are selected through a mini-PointNet \cite{pointnet++}, in which $N$ key proposal points are sampled using farther point sampling (FPS) \cite{pointnet}, and the corresponding $N$-bunch semantic features are further enhanced. Finally, the enhanced $N$ proposals are fed into a 3-layer convolutional task layer to output 5-dimensional vectors, including 1-dimensional proposal-wise scores, 3-dimensional refined offsets and 1-dimensional orientations.

In contrast, we propose a simple, effective and interpretable one-stage point-to-box network. On the one hand, a parallel layer composed of 2-layer 1D convolution is designed to learn task-interactive features to explicitly align accuracy, as well as a task-specific layer is designed to learn task-specific features to predict different task parameters synchronously. On the other hand, we also devise a classification layer to reason foreground-background classification scores. Leveraging the classification scores, we adjust the center-ness scores to output the proposal-wise scores.

\subsection*{C. Performance Comparison}
\label{subsection3}
All experiments are conducted on Car categories from KITTI \cite{kitti} dataset using office code\textcolor{red}{\footnote{\textcolor{red}{\url{https://github.com/HaozheQi/P2B}}}}. Here, we use P2B-ours to represent the tracking method, which replaces the two-stage point-to-box network in P2B with our one-stage point-to-point network.

\begin{figure}[t]
    \centering
    \includegraphics[width=1.6in]{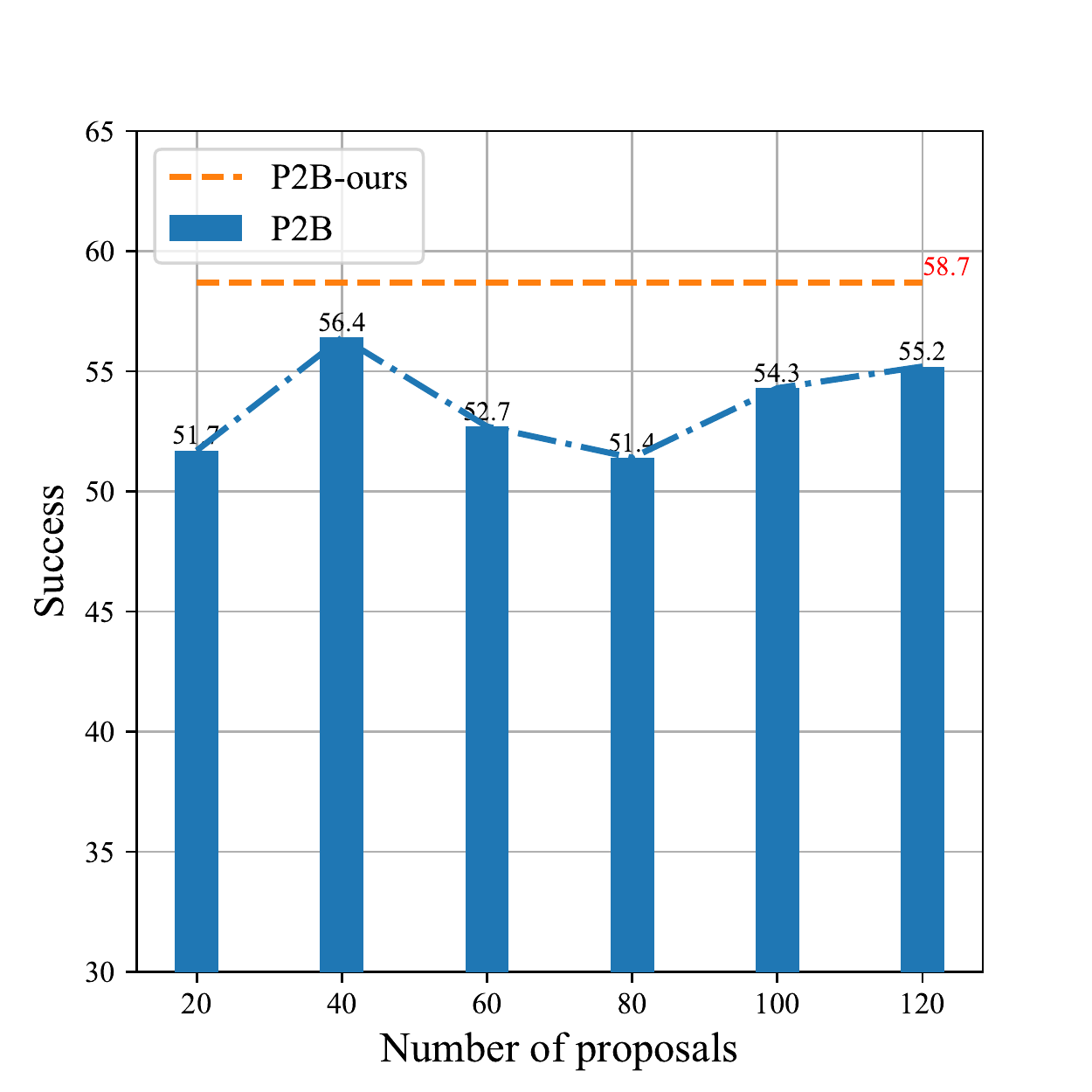}
    \includegraphics[width=1.6in]{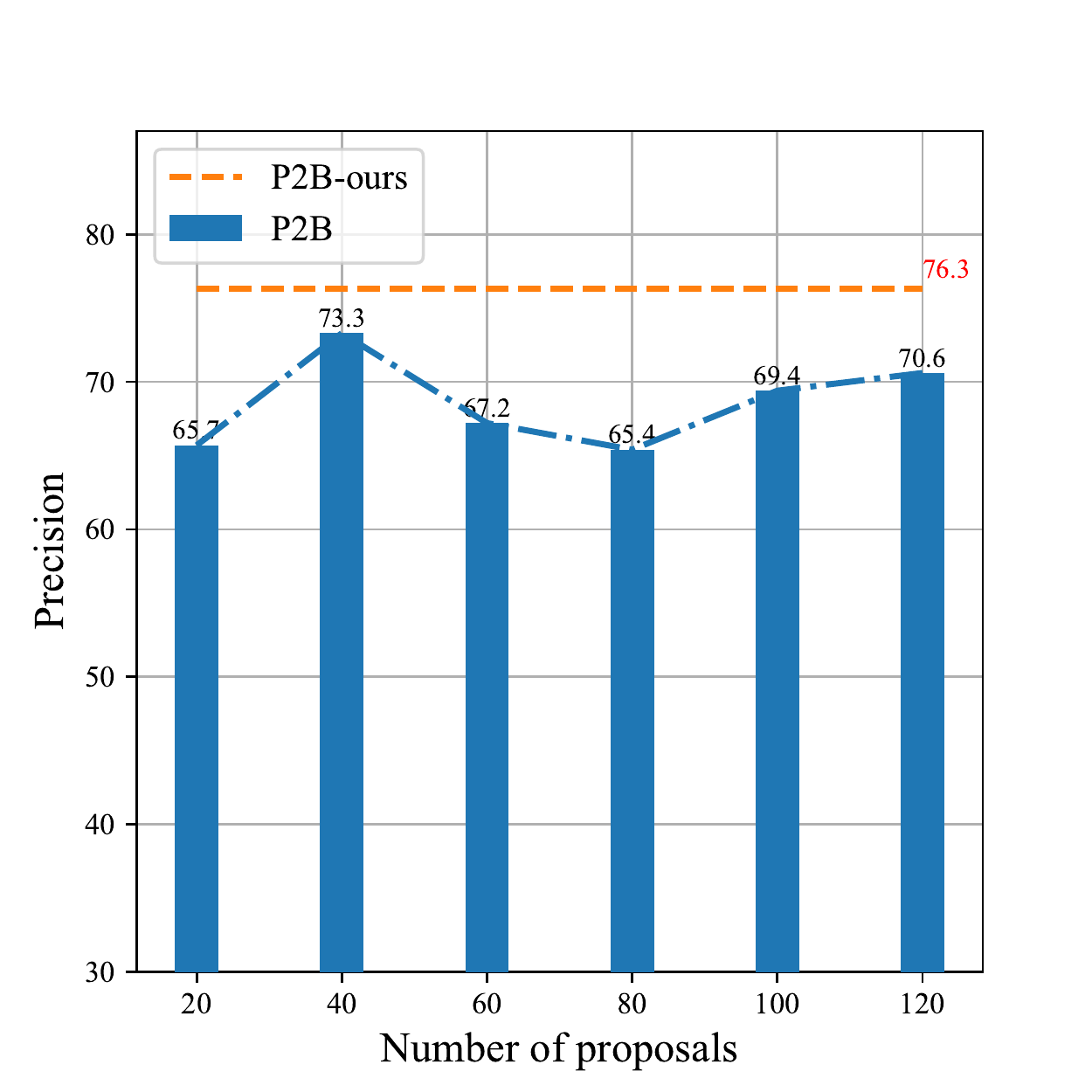}
    \caption{Performance comparison under different number of proposals. The left sub-figure and right sub-figure are plotted in terms of \textit{Success} and \textit{Precision}, respectively.}
    \label{fig8}
\end{figure}

\noindent\textbf{C.1. Tracking-Sensitive Parameters. }In this paper, we propose a one-stage point-to-box network that gets rids of some tracking-sensitive parameters, such as the number of proposals, a crucial hyper-parameter existing in the two-stage point-to-box network. A large number will introduce more distracting information, while a small number will result in the optimal proposals being discarded. Therefore, P2B method is sensitive to it and exhibits widely varying performance under the different number of proposals, as shown in Fig. \ref{fig8}. Thanks to our one-stage design, P2B-ours method is free of this parameter and achieves the better \textit{Success} and \textit{Precision} in values of 58.7\% and 76.3\%.

\begin{figure}[t]
    \centering
    \includegraphics[width=1.6in]{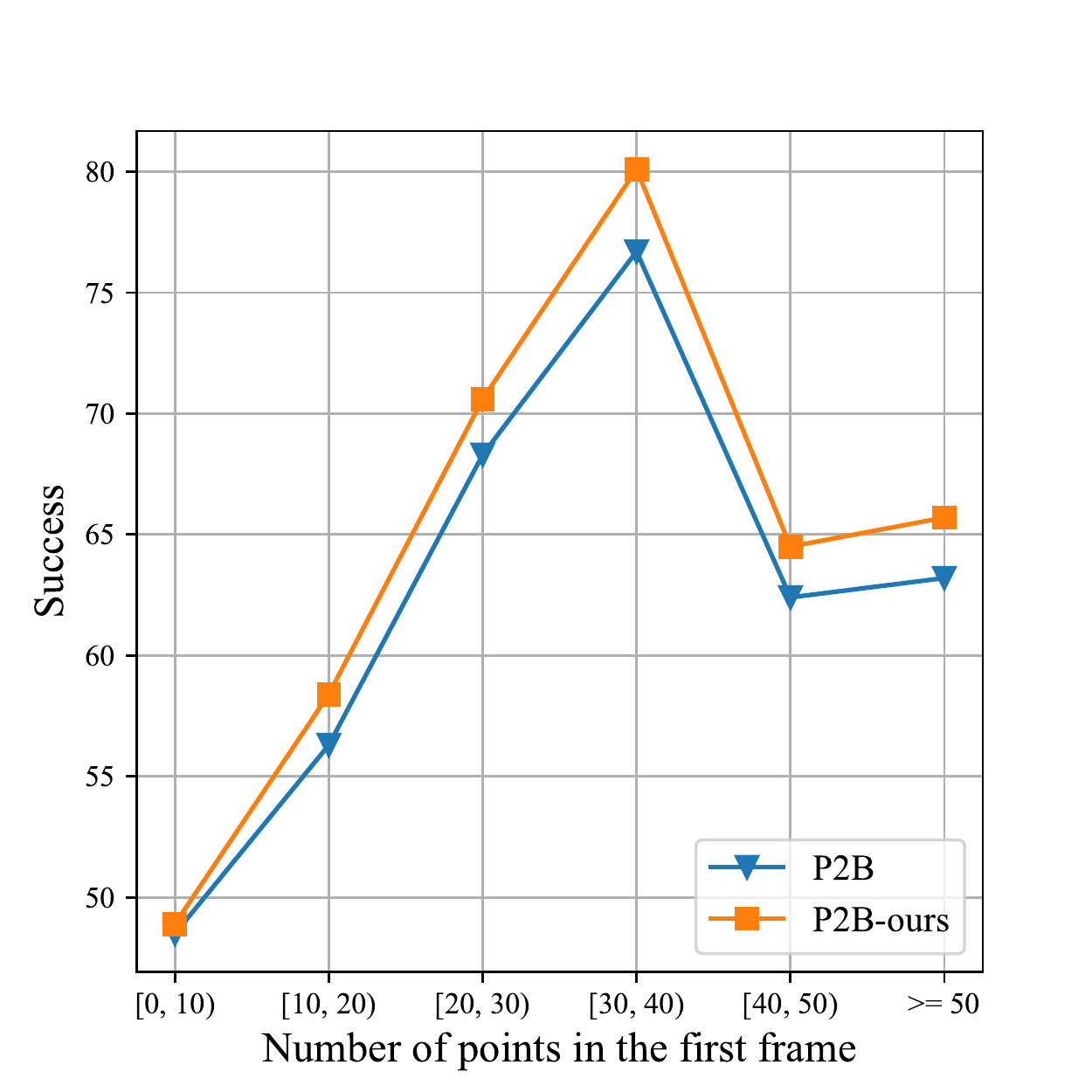}
    \includegraphics[width=1.6in]{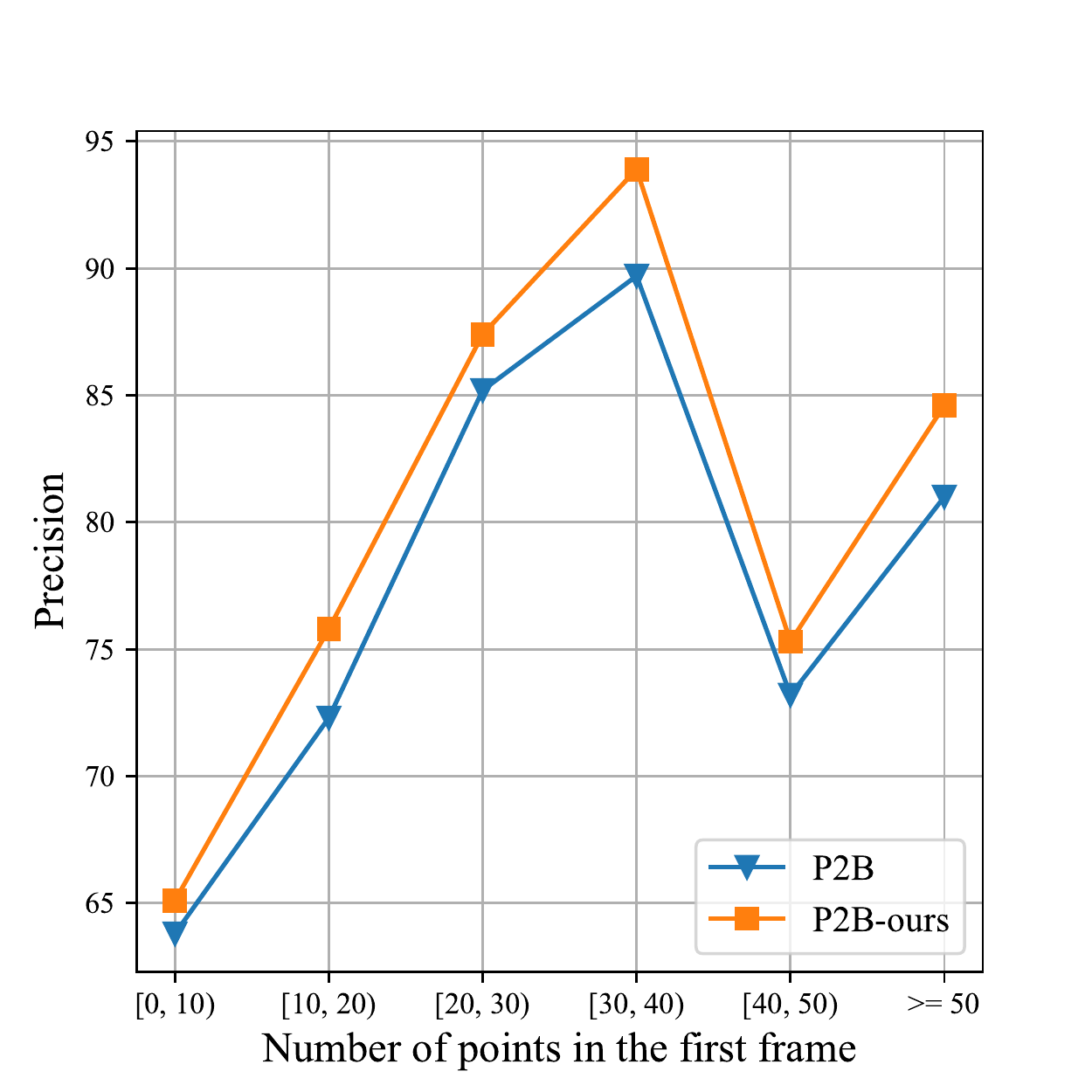}
    \caption{Performance comparison on different sparsity levels. The left sub-figure and right sub-figure are plotted in terms of \textit{Success} and \textit{Precision}, respectively.}
    \label{fig9}
\end{figure}

\noindent\textbf{C.2. Robustness to Sparse Point Clouds. }In fact, point clouds captured by LiDAR sensors are usually sparse and incomplete. Thus, it is a practical requirement for trackers to analyze their robustness to sparse point clouds. For any point cloud sequence, the first frame with less than 50 points is defined as a sparse scene, and about 76\% (4870/6424) of the total number of frames satisfy this condition. According to sparsity level, these frames are further divided into five kinds: [0,10), [10,20), [20,30), [30,40) and [40,50) with corresponding 2394, 1590, 709, 97 and 80 frames, respectively. As shown in Fig. \ref{fig9}, P2B-ours shows better robustness to sparse point clouds than P2B, which is contributed to the proposed one-stage point-to-box network.

\begin{figure*}[t]
    \centering
    \includegraphics[width=1.9\columnwidth]{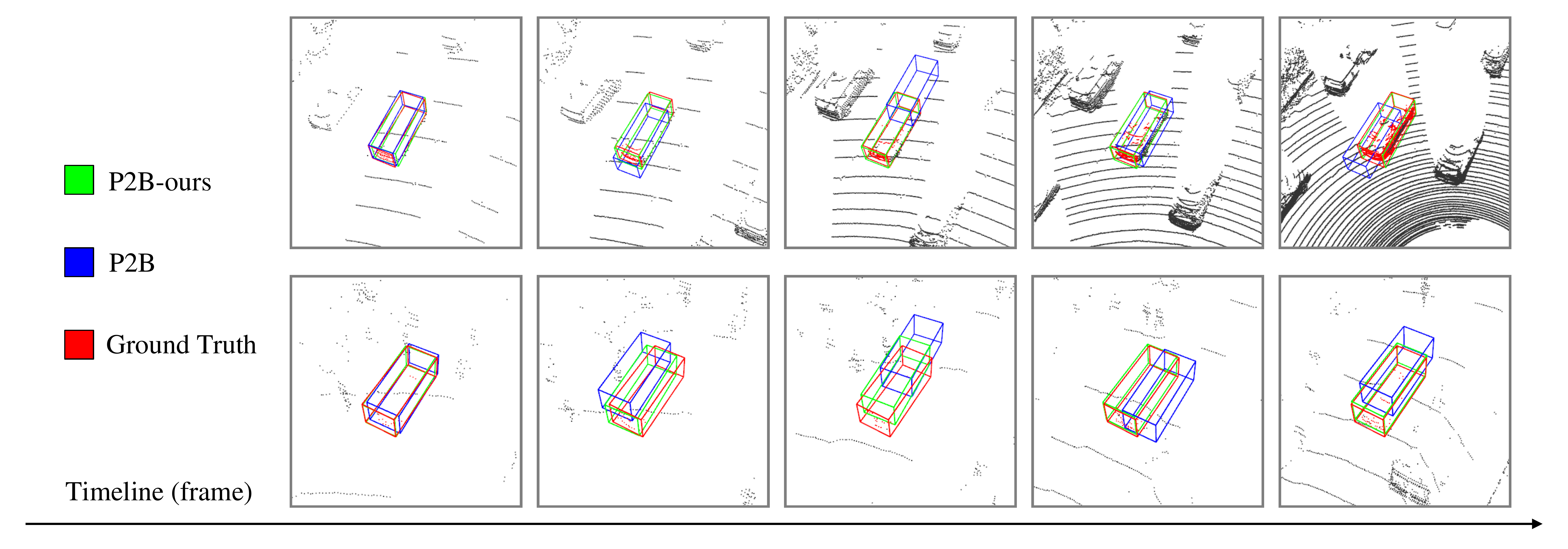}
    \caption{Advantageous cases of our P2B-ours over P2B in both dense (top row) and sparse (bottom row) point cloud sequences.}
    \label{fig10}
\end{figure*}

\begin{table*}[t]
    \centering
    \begin{tabular}{cccccc}
      \toprule
      Source of template & The First GT & Previous result &  All previous rersults & \underline{First $\&$ Previous}\\
      \midrule
      \midrule
      P2B \cite{p2b}& 46.7 / 59.7 & 53.1 / 68.9 & 51.4 / 66.8 & 56.2 / 72.8 \\
      P2B-ours & \textbf{47.8} / \textbf{61.3} & \textbf{55.7} / \textbf{72.4} & \textbf{53.3} / \textbf{68.7} & \textbf{58.7} / \textbf{76.3} \\
      \bottomrule
    \end{tabular}
    \caption{Comparison results of different ways for template generation. ``\underline{First $\&$ Previous}'' represents ``the first ground $\&$ previous result'', which is the default way used in our method. \textbf{Bold} denotes the best result.}
    \label{table6}
\end{table*}

\begin{table}[t]
    \centering
    \begin{tabular}{ccc}
      \toprule
      Method & \underline{Long Term} & Short Term\\
      \midrule
      \midrule
      P2B \cite{p2b}& 56.2 / 72.8 & 81.2 / 88.5\\
      P2B-ours & \textbf{58.7} / \textbf{76.3} & \textbf{81.9} / \textbf{88.7}\\
      \bottomrule
    \end{tabular}
    \caption{Comparison results of long term and short term ways of search region generation. \textbf{Bold} denotes the best performance. ``\underline{Long Term}'' is the default way used in our method.}
    \label{table7}
\end{table}
\noindent\textbf{C.3. Ways for Template Region Generation. }As default experiment setting, the template region of $t$-th frame is formed by merging the points inside the ground truth BBox of 1-st frame and previously predicted BBox of $(t-1)$-th frame. Here, to further prove our work, we investigate the impact of different ways for template generation (i.e., ``the first ground truth'', ``the previous result'' and ``all previous results'') on tracking performance. The results are reported in Table \ref{table6}. P2B-ours methods outperforms P2B in all ways by a significant margin. Especially in the case of ``the previous result'', 2.6 \% and 3.5 \% improvements in terms of Success and Precision, respectively are gained due to the proposed point-to-box network. This means that our method brings less template information errors, and therefore performs better tracking in the case where only the previously predicted BBox is used to form the template region.

\noindent\textbf{C.4. Ways for Search Region Generation. }In addition, we further explore different ways to generate search regions. For long term tracking, we generate the current search region of $t$-th frame using the previously predicted BBox of $(t-1)$-th frame. In this way, trackers need to deal with the accumulating tracking errors from previous frames. While for short term tracking, we utilize the ground truth BBox of $(t-1)$-th frame as the base region to generate the current search region. With this way, tracking task is defined as ``just-in-time tracking''. Although the previous ground truth BBoxes are not available during inference, it can be a fair reflection of the tracker's accuracy. We report the results in Table \ref{table7}. Our method performs better than P2B is both two ways. For short term tracking style, the more advanced performance of P2B-ours straightforwardly demonstrates that our method can guided more accurate tracking results.

\noindent\textbf{C.5. Quantitative Results. }In Fig. \ref{fig10}, we visualize the advantage of one-stage point-to-box network based P2B-ours over the two-stage point-to-box network based P2B. We plot a dense point cloud sequence and a sparse point cloud sequence. It can be clearly seen that our methods predicts tracking BBoxes more accurately than P2B.

\noindent\textbf{C.6. Model Parameters, Flops and Speed. }Here, we compare the parameters, floating point operations per second (Flops) and tracking speed in Table \ref{table8}. Note that, the parameters and Flops are only counted in prediction head, i.e., one- and two-stage point-to-box network. As reported, our one-stage model enjoys considerably fewer parameters and flops than the two-stage model, and thus runs 7 Fps faster.
\begin{table}[t]
    \centering
    \begin{tabular}{cccc}
      \toprule
      & Parameters & Flops  & Speed \\
      \midrule
      \midrule
      P2B \cite{p2b}& 0.7$\times 10^6$& 2.4$\times 10^9$ & 28 Fps \\
      P2B-ours &\textbf{0.3$\times 10^6$} &\textbf{0.3$\times 10^9$}  & \textbf{35 Fps}\\
      \bottomrule
    \end{tabular}
    \caption{Comparison results of flops and speed. \textbf{Bold} denotes the best performance.}
    \label{table8}
\end{table}

\end{document}